# Looking for plausibility

**Wan Ahmad Tajuddin Wan Abdullah**
*Faculty of Computer Science and Information Technology, Universiti Malaya, 50603 Kuala Lumpur*
and
*Department of Physics, Faculty of Science, Universiti Malaya, 50603 Kuala Lumpur*

Abstract:

In the interpretation of experimental data, one is actually looking for plausible explanations. We look for a measure of plausibility, with which we can compare different possible explanations, and which can be combined when there are different sets of data. This is contrasted to the conventional measure for probabilities as well as to the proposed measure of possibilities. We define what characteristics this measure of plausibility should have.

In getting to the conception of this measure, we explore the relation of plausibility to abductive reasoning, and to Bayesian probabilities. We also compare with the Dempster-Schaefer theory of evidence, which also has its own definition for plausibility. Abduction can be associated with biconditionality in inference rules, and this provides a platform to relate to the Collins-Michalski theory of plausibility. Finally, using a formalism for wiring logic onto Hopfield neural networks, we ask if this is relevant in obtaining this measure.

INTRODUCTION

Traditionally, uncertainty in propositions is handled through the concept of probability with its statistical interpretation. It is however, a bit strange when such an approach which is based on counting events, is applied to hypotheses, like explanatory propositions. For example, in the interpretation of experimental data, one is actually looking for plausible explanations to describe the observed data. Whilst probability may be seen to underlie data measurement, a statistical picture for explanations may be viable only in a multiple-world universe.

We thus seek a measure for plausibility, with which we can compare different possible explanations. Furthermore, this measure of plausibility should allow appropriate combination when different explanations and/or sets of data are combined.

UNCERTAINTY MEASURES

Probabilities are calculated from the numbers of events satisfying respective propositions from a total set and are normalized. Thus probabilities for alternative explanations are exclusive, and the more alternatives there are, the less are the values for probabilities in general. Plausibility, on the other hand, should not behave as such.. The plausibility of a proposition should not depend on the plausibility of an alternative proposition, much less reduced by it.

In logical conjunctions, probabilities are combined by multiplication, and in disjunctions, by addition, when the atoms are exclusive:
$$P(X \wedge Y) = P(X) \times P(Y)$$
$$P(X \vee Y) = P(X) + P(Y) - P(X \wedge Y)$$
This does not necessarily follow for plausibilities; in fact, for example, intuitively it does not seem correct that the plausibility of ($X$ or $Y$) is the sum of the individual plausibilities of $X$ and $Y$.

Plausibility seems closer to the notion of possibility [1] (based on fuzzy sets, thus rather similar to fuzzy logic [2]) when combinations are concerned. Disjunctions of possibilities return the maximum values, while conjunctions of possibilities involve the minimum values:
$$pos(X \wedge Y) \leq min(pos(X), pos(Y))$$
$$pos(X \vee Y) = max(pos(X), pos(Y))$$

Other than probabilities and possibilities, other uncertainty measures (sometimes with similar names) have been proposed by, for example Dempster and Shafer [3], further studied below, and others.

PLAUSIBILITY

Let us now define what characteristics the measure of plausibility should have.

- We can have a normalized scale from 0, which would denote something not at all plausible, to 1 denoting something fully plausible. Let us denote plausibility by '$pl$':
$$pl(X) = 0 : X \text{ is not all plausible}$$
$$pl(X) = 1 : X \text{ is fully plausible}$$

- Plausibilities are not exclusive. For example, even when $X$ and $Y$ are not logically compatible, yet $X$ can be fully plausible at the same time that $Y$ is fully plausible:
$$pl(X) + pl(Y) > 1 \text{ though } X \cap Y = \emptyset$$

- Like possibility, and unlike probability, plausibility is not self-dual. That is,
$$pl(X) \neq 1 - pl(\neg X).$$
In another words, when something is not at all plausible, it does not mean that it is fully plausible that that something is untrue. Let us include negative values for plausibility to enable the representation of negations, and require that
$$pl(\neg X) = - pl(X).$$
This means that now
$$pl(X) \in [-1, 1],$$
giving values for fully plausibly false on the one end, and fully plausibly true on the other, with non-plausibility ('non-relevance') in the middle.

A Wiktionary definition of plausibility is "seemingly or apparently valid, likely, or acceptable". This suggests that plausibility may be related to abduction.

## ABDUCTIVE REASONING

Abduction (see e.g. [4]), first described by Peirce more than a hundred years ago, is the process of arriving at the premise which would 'explain' some situation. Given that the set of propositions $C$ formally follows from the set of propositions $A$ subject to the set of logical rules $T$, then the derivation of $C$ from $T$ and $A$ is deduction, that of $T$ from $C$ and $A$ is induction, and that of $A$ from $T$ and $C$ is abduction. Note that the abduced set is sufficient, but not necessary for $C$ to follow from $T$, and may be one of many alternatives; in abduction one also usually looks for the most natural explanation in the form of the most economical one.

Given a set of true propositions, then, an abduced proposition is a plausible proposition. Plausibility then is a measure of how good is the proposition in explaining the available facts. It can be related to how much of the requirement for sufficiency has been satisfied.

## BAYESIAN PROBABILITIES

When propositions have probabilistic values, then abduction is related to conditional probabilities. Bayes' theorem states that the probability for $\theta$ given $x$,
$$P(\theta|x) = P(x|\theta) P(\theta) / P(x)$$
depends on the conditional probability, or likelihood, of $x$ given $\theta$, multiplied by the prior probability for $\theta$. Abduction being the 'reverse' of deduction is parallel to the conditional probability being the 'reverse' of likelihood.

Conditional probabilities then present a possible model for plausibilities. Combinations of plausibilities in inferences can copy those in Bayesian networks [5]. However, the determination of prior probabilities pose a problem for calculating conditional probabilities.

## DEMPSTER-SHAFER THEORY

The Dempster-Shafer theory of evidence [3] has its own definition for plausibility which is based on some belief function. Plausibility for a set of propositions is the sum of "belief masses" of other sets of propositions which intersect with this set, while belief is that for those which are subsets, and they bound the probability:
$$bel(X) \leq P(X) \leq pl(X)$$
and are related,
$$pl(X) = 1 - bel(\neg X).$$
The theory prescribes a method for combining belief masses from different assignments, along the lines of Bayesian theory. To obtain the plausibility of a combined set of propositions, one has to recalculate using the primary definition.

The assignment of belief masses is problematic here, and also, their association with probabilities is misleading. Furthermore, plausibility being the upper bound for belief suggests a rather subjective definition.

# BICONDITIONALITY AND COLLINS-MICHALSKI

A theory for plausible reasoning has been proposed by Collins and Michalski [6]. It has several types of inferences, namely mutual implication, mutual dependency, generalization, specialization, and similarity/analogy. A drawback is that it has a restrictively large number of parameters to model uncertainty.

The inferences existing in Collins-Michalski theory are mostly explained by biconditionality, wherein implications are taken to also include some degree of equivalences [7]. This relates to abduction, as abduction coincides with deduction when implications are replaced by equivalences. In the probabilistic picture, this is like equating the conditional probability to the likelihood,
$$P(\theta|x) = P(x|\theta)$$
Of course this is not rational; after all, this is plausible reasoning. However, this supports the notion that plausibility has an abductive basis.

# CONNECTIONIST

It has been shown [8,9] that logic can be hardwired onto a Hopfield neural network [10]. The Hopfield network minimizes a Lyapunov or "energy" function related to the synaptic strengths [11]. By writing an expression yielding a value proportional to unsatisfied clauses, the network then searches for an optimum logical interpretation of the propositional atoms, represented by neurons. This then gives the values for the synapses. In some sense, the logic is contained in the synapses.

From the synaptic strengths then, we can make inferences about the logical clauses [12]. This may be helpful in our search for a measure of plausibility.

# PLAUSIBILITY, PLAUSIBLY

Finally, we can make some preliminary moves in the direction of plausibility. Taking an abductive interpretation, the more of the set of existing 'measured' propositions $C$ a hypothetical proposition $A_i$ forces, the more plausible $A_i$ should be. Perhaps then the plausibility of $A_i$ corresponds to the fraction of propositions in $C$ that is forced correct (both, either true or false) by $A_i$. If $A_i$ does not have any effect on $C$ then $A_i$ should have plausibility 0, and if forces all of the known propositions $C$ to be correct, then $A_i$ have plausibility 1.

What is of interest is the combination of plausibilities. Is it possible to have simple combining rules for conjunctions and disjunctions, for example? For propositions $A_i$ and $A_j$ each forcing disjoint subsets of C and having no joint effects on the correctness of any subset, then simply
$$pl(A_i \wedge A_j) = pl(A_i) + pl(A_j).$$
In the case of the existence of some common proposition $C_i$ forced by both $A_i$ and $A_j$, i.e. the rule base $T$ contains
$$A_i \to C_i$$
$$A_j \to C_i$$
then the value is less, as also implicated by the upper limit of 1 on the value of plausibilities, while in the case that $A_i$ and $A_j$ have a joint effect on $C_i$, e.g.
$$A_i \wedge A_j \to C_i$$

then the value is more. In general, plausibility of conjunctions can be complicated. It may simplify somewhat for singleton *C*'s, whence the measure would involve some kind of probability, and these phenomena would be averaged out.

Having plausibility related to fraction of forced correctness is compatible with our previous definition of negative plausibilities. For disjoint hypotheses $A_i$ and $A_j$, using de Morgan's theorem, the plausibility for a disjunction is similarly given,

$$pl(A_i \vee A_j) = pl(A_i) + pl(A_j),$$

a bit counter-intuitively. Proceeding, thus

$$pl(A_i \rightarrow A_j) = pl(A_j) - pl(A_i).$$

Interestingly, this has some similarity to synaptic weight assignment in connectionist logic previously discussed where whilst some weights are increased with the addition of a clause, some are decreased.

Using hardwired logic on connectionist networks, and forcing the correct values for the neurons in *C*, and that for a hypothesis, the final network state yields the least value for the energy function, which corresponds to the number of clauses unsatisfied. This intuitively should also represent the number of errors in *C*, allowing an estimation of the plausibility of the hypothesis. For this, synaptic weights, or equivalently, clauses in *T* need to be known explicitly; however, this can be obtained [12] through Hebbian learning.

CONCLUSIONS

In summary, we have explored the concept of plausibility and attempted to characterize a measure for it. In particular, we have taken an abductive basis, and conjecture that a hypothesis is more plausible if it forces more of the measured truth. For simple cases we can write down the rules for combinations. We also indicated how plausibility can be measured using a logic neural network.

This exploratory proposal for plausibility needs to be further investigated, especially for the non-trivial cases of combinations. It is hoped that combination of plausibilities would help in the interpretation of combinations of measurements in experimental data.